\def\year{2021}\relax
\documentclass[letterpaper]{article} 
\usepackage{aaai21}  
\usepackage{times}  
\usepackage{helvet} 
\usepackage{courier}  
\usepackage[hyphens]{url}  
\usepackage{graphicx} 
\usepackage{algorithm}
\usepackage{algorithmic}
\usepackage{longtable}
\usepackage{caption}
\usepackage{subcaption}
\usepackage{amsmath}
\usepackage{amsfonts}
\usepackage{booktabs}       
\usepackage{supertabular}
\usepackage{multicol}

\usepackage{natbib}  
\usepackage{caption} 
\nocopyright
\title{A Cross-Level Information Transmission Network for Predicting Phenotype from New Genotype: Application to Cancer Precision Medicine}

\author {
        Di He,\textsuperscript{\rm 1}
        Lei Xie \textsuperscript{\rm 2} \\
}
\affiliations {
    \textsuperscript{\rm 1} PhD Program in Computer Science, 
    Graduate Center City University of New York\\
    \textsuperscript{\rm 2} Department of Computer Science,
    Hunter College, City University of New York \\
    dhe@gradcenter.cuny.edu, lxie@iscb.org
}
\begin{document}

\maketitle

\begin{abstract}
An unsolved fundamental problem in biology and ecology is to predict observable traits (phenotypes) from a new genetic constitution (genotype) of an organism under environmental perturbations (e.g., drug treatment). The emergence of multiple omics data provides new opportunities but imposes great challenges in the predictive modeling of genotype-phenotype associations. Firstly, the high-dimensionality of genomics data and the lack of labeled data often make the existing supervised learning techniques less successful. Secondly, it is a challenging task to integrate heterogeneous omics data from different resources. Finally, the information transmission from DNA to phenotype involves multiple intermediate levels of RNA, protein, metabolite, etc. The higher-level features (e.g., gene expression) usually have stronger discriminative power than the lower level features (e.g., somatic mutation). To address above issues, we proposed a novel Cross-LEvel Information Transmission network (CLEIT) framework. CLEIT aims to explicitly model the asymmetrical multi-level organization of the biological system. Inspired by domain adaptation, CLEIT first learns the latent representation of high-level domain then uses it as ground-truth embedding to improve the representation learning of the low-level domain in the form of contrastive loss. In addition, we adopt a pre-training-fine-tuning approach to leveraging the unlabeled heterogeneous omics data to improve the generalizability of CLEIT. We demonstrate the effectiveness and performance boost of CLEIT in predicting anti-cancer drug sensitivity from somatic mutations via the assistance of gene expressions when compared with state-of-the-art methods.
\end{abstract}

\section{Introduction}
Advances in next-generation sequencing have generated abundant and diverse omics data. They provide us with unparalleled opportunities to reveal the secrets of biology. An unsolved problem in biology is how to predict observable traits (phenotypes) given a new genetic constitution (genotype) under environmental perturbations. The predictive modeling of genotype-phenotype associations will answer not only many fundamental questions in biology but also address urgent needs in biomedicine. A typical application is anti-cancer precision medicine. Given a new cancer patient's genetic information, what is the best existing drug to treat this patient? This is different from Genome Wide Association Study (GWAS) and Transcriptome Wide Association Study (TWAS), whose goal is to identify statistical correlations between observed genotype and phenotype. The predicting phenotype from a new genotype is a challenging task due to the asymmetrical multi-level hierarchical organization of the biological system. Cell-, tissue-, and organism-level phenotypes do not arise directly from DNAs but through multiple intermediate molecular or cellular phenotypes that are characterized by biological pathways, protein interactions, and gene expressions, etc \cite{Blois1984Scalar}. In other words, in the information transmission process from DNA to RNA to protein to the observed phenotype of interest, the higher-level features (e.g., gene expression) usually have stronger discriminative power than the lower level features (e.g., somatic mutation) in a supervised learning task that is independent on the machine learning model applied. This premise is supported by multiple studies such as cancer \cite{costello2014community}, drug combination \cite{menden2019community}, and microbiome \cite{Lloyd2019Microbiome}. Therefore, a multi-level approach is needed to simulate the asymmetrical hierarchical information transmission process for linking the genotype to the phenotype \cite{hart2016providing}. Furthermore, the interpretability of machine learning model is critical for the biomedical application. The multi-scale modeling of genotype-phenotype associations will facilitate opening the black box of machine learning \cite{Yang2019WhiteBox}. In addition to the above fundamental challenge, the predictive modeling of genotype-phenotype associations faces several technical difficulties that hinder the application of existing machine learning methods. Firstly, omics data are often in an extremely high dimension. Secondly, the labeled data are scarce compared with unlabeled data. Finally, it is not a trivial task to integrate heterogeneous omics data from different resources. 

To address the aforementioned challenges, we develop a novel neural network-based framework: Cross-LEvel Information Transmission (CLEIT) network. Inspired by domain adaptation, CLEIT first learns the latent representation of a high-level domain then uses it as ground-truth embedding to improve the representation learning of the low-level domain in the form of contrastive loss. In addition, we adopt a pre-training-fine-tuning approach to leveraging the unlabeled heterogeneous omics data to improve the generalizability of CLEIT. We demonstrate that CLEIT is effective in predicting anti-cancer drug sensitivity from somatic mutations, and significantly outperforms other state-of-the-art methods. Precision anti-cancer therapy that is tailed to individual patients based on their genetic profile has gained tremendous interest in clinical \cite{precisionmed}. Cancer acquires numerous mutations during its somatic evolution. Both driver and passenger mutations collectively confer cancer phenotypes and are associated with drug responses \cite{passenger_mutation}. Thus it is necessary to use the entire mutation profile of cancer for the prediction of anti-cancer drug sensitivity. The machine learning models that can explicitly model hierarchical biological processes will no doubt facilitate the development of precision medicine.   

\section{Related Work}

CLEIT borrowed some ideas from widely used domain adaptation techniques. Domain adaptation aims at transferring the knowledge a trained predictive model has gained on the source domain with sufficient labeled data to the target domain without or with limited labeled data when the source and target domains are of different data distributions. In particular, feature-based domain adaptation approaches \cite{weiss2016survey} have gained popularity along with the advancement in deep learning techniques due to its power in feature representation learning. It aims to learn a shared feature representation by minimizing the discrepancy across different domains while leveraging supervised loss from labeled source examples to maintain trait space's discriminative power. To achieve discrepancy reduction, there are typically two main methodologies.

The first one focuses on exploring proper statistical distribution discrepancy metrics. For example, maximum mean discrepancy (MMD) \cite{MMD} is used in deep domain confusion (DDC) \cite{DCC2014} as domain confusion loss on the domain adaptation layer to learn domain invariant features in addition to the regular classification task. Deep adaptation network (DAN) \cite{DAN2015}, and its follow-up works \cite{rtn, jan} explored the idea of using multi-kernel MMD to match mean embedding of the multi-layer representations across domain and enhanced the feature transferability. In addition to MMD, CORAL \cite{coral2015} matches the data distributions with second-order statistics (co-variance) on linear transformed inputs,  and is further developed into its non-linear variant in Deep CORAL \cite{deepcoral2016}. Furthermore, Wasserstein distance is employed in JDOT \cite{jdot} and WGDRL \cite{wdgrl}.

The other scheme is inspired by domain adaptation theory \cite{ben2007analysis, ben2010theory}. It states that predictions must be made based on features that can not discriminate between source and target domains to achieve effective domain adaptation transfer. It intends to minimize the distribution difference across domains by adopting an adversarial objective with a trainable domain discriminator. Domain adversarial neural network (DANN)  \cite{dann2016} learned domain invariant features by a minimax game between the domain classifier and the feature generator with layer sharing and customized gradient reversal layer. Later, adversarial discriminative domain adaptation (ADDA) \cite{adda2017} achieved domain adaptation via a general framework consisting of discriminative modeling, untied weight sharing, and a GAN loss. ADDA first performs the discriminative task with labeled source domain samples and then utilizes the GAN \cite{gan} architecture to learn the mapping from target domain samples to source domain feature space.

In addition to the above-mentioned discrepancy reduction approaches, encoder-decoder models are also widely used in domain adaptation, where domain invariant features are learned via shared intermediate representation while domain-specific features are preserved with reconstruction loss. Representative works include marginalized denoising autoencoder \cite{mDAE}, multi-task autoencoders \cite{mtAE}, deep reconstruction classification network \cite{drcn}. Moreover, domain separation network (DSN) \cite{dsn} was proposed to explicitly separate private representations for each domain and shared representations across domains. The shared representation is learned similarly as DANN \cite{dann2016} or with MMD \cite{MMD}, while the private representations are learned via orthogonality constraint against the shared representation. DSN can achieve better generalization across domains with the reconstruction through the concatenation of shared and private representations than other methods. 

\section{Contributions}
CLEIT aims to address an important problem of multi-scale modeling of genotype-phenotype associations. Although CLEIT borrowed some ideas from the domain adaptive transfer learning, there is a significant difference between those approaches and CLEIT. The goal of classic domain adaptation is to use the label information from the source domain data to boost the performance of supervised tasks in the target domain without abundant labels. The feature in the target domain usually has a similar discriminative power to that in the source domain. While in our case, we focus on resolving the inherent discriminative power discrepancy between two domains, which have a hierarchy relation. The feature of the high-level domain has higher discriminative power than that of the low-level domain. Moreover, the entity types of source and target domains are usually the same in conventional domain adaptation. In our case, they are of different types. Specifically, our goal for information transmission is to solely push the latent representation of the low-level domain to approximate the one of the high-level domain, that is, the feature representation learned from the high-level domain is fixed and used as ground-truth feature representation of the low-level domain. In this setting, the latent space where the cross-level information transmission happened is no longer a symmetrical consensus from different domains. To boost the discrimination power of the low-level domain, the high-level and low-level domain is used as an input and an output, respectively. A mapping function is learned between them. 

The major contributions of this research are summarized as follows.

- We design a pre-training-fine-tuning strategy to fully utilize both labeled and unlabeled omics data that are naturally noisy, high-dimensional, heterogeneous, and sparse. 

- We propose a novel neural network framework that can explicitly model asymmetrical cross-level information transmissions in a complex system to boost the discriminative power of the low-level domain. The multi-level hierarchical structure is the fundamental characteristic of the biological and ecological system. The proposed architecture is general and can be applied to model various machine learning tasks in a multi-level system. 

- In terms of biomedical application, the CLEIT model significantly improves individualized anti-cancer sensitivity prediction using only somatic mutation data. The oncology panel of somatic mutations has been routinely performed in the cancer treatment. The application of CLEIT may improve the effectiveness of cancer treatment and achieve precision medicine. 
\section{Method}
\label{Methods}
\subsection{Problem formulation}
We denote a data domain \(D \) as \(D=\{\mathcal{X}, P(X)\}\), where \(\mathcal{X} \) stands for the feature space and samples within domain \(D\), \(X=\{x_1, \dots, x_n\}\in \mathcal{X}\). \(P(X)\) is the affiliated marginal distribution. 
In this work, we consider two domains \(D_H=\{\mathcal{X}_h, P_h(X_h)\}\) and \(D_L=\{\mathcal{X}_l, P_l(X_l)\}\), namely the high-level domain and low-level domain, where \(\mathcal{X}_h \neq \mathcal{X}_l \), \(P_h(X_h) \neq P_l(X_l)\). In addition,  one common task \(\tau\) of interest is to predict phenotype or other outcomes. This task can be done individually from both of the domains but with different performance, where \(D_H\) can achieve superior performance to \(D_L\) independent on machine learning models applied to them. Here, the performance difference is due to the nature of each domain's data, instead of the volume of labeled samples as in a classical domain adaptation setting. However, although feature space \(\mathcal{X}_h\) and \( \mathcal{X}_l \) are not the same, the entities cross the feature spaces are hierarchically related, such as the multi-level hierarchical organization of omics data of a biological system. Based on this realization, the aim is to utilize the knowledge learned from \(D_H\) to boost the predictive power of \(D_L\).

\subsection{CLEIT framework}
To use the knowledge learned from \(D_H\) to boost the performance of \(D_L\), we propose a Cross-LEvel-Information Transmission (CLEIT) framework. The strategy of CLEIT is to encode the data from both domains into certain "higher-level" features. The embedded "high-level" feature has the direct implication of the task of interests and achieves the cross-level information transmission through transferring knowledge via learned representations cross domains. 

\begin{figure*}[t!]
\centering
\includegraphics[width=1.0\textwidth]{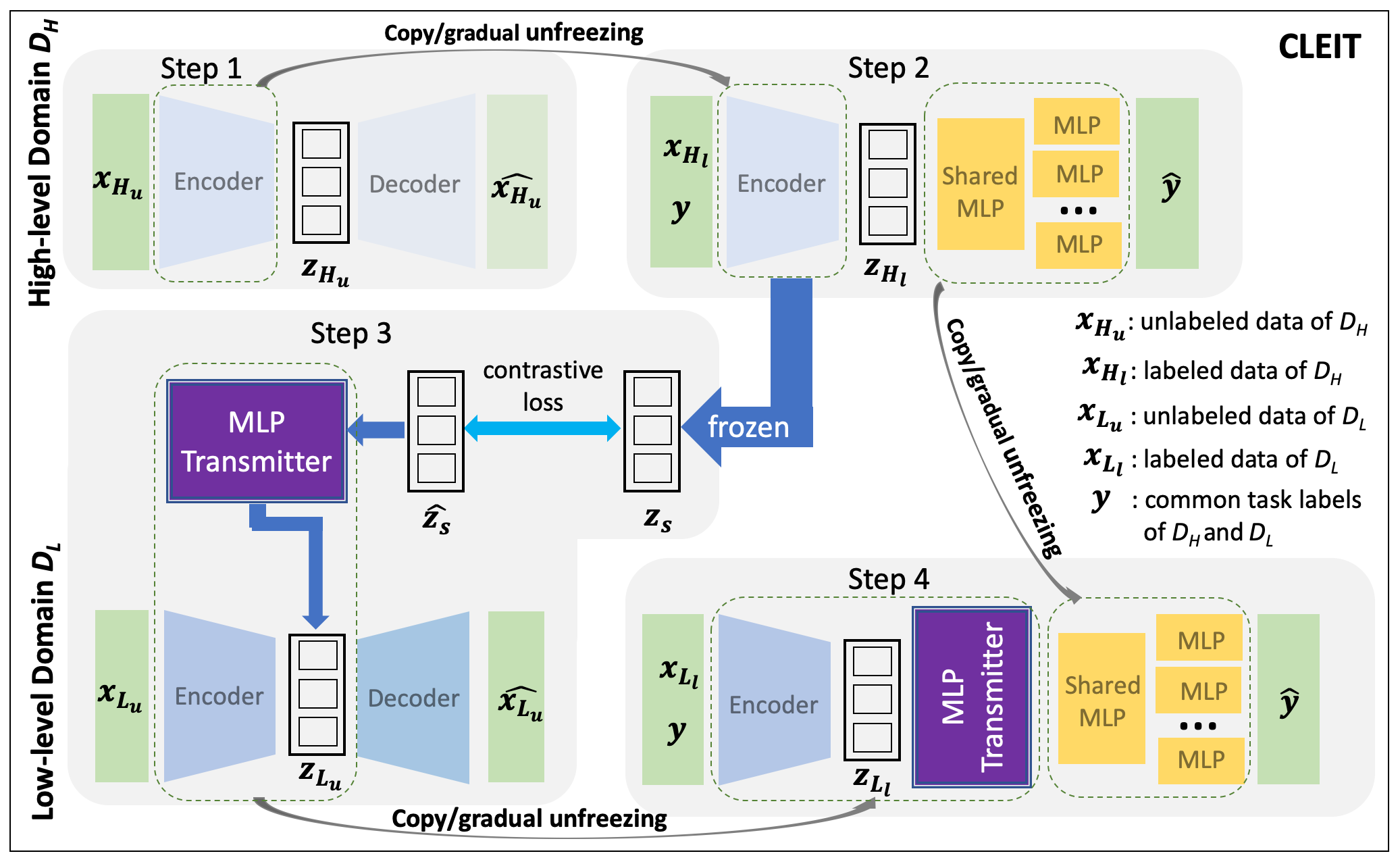}
\caption{CLEIT Framework. The training of CLEIT involves four steps. First, the encoder of \(D_H\) is learned from an autoencoder, and fine-tuned by a supervised multi-task MLP in the step 1 and 2. Then, the embedding of \(D_L\) is encoded from an autoencoder, and the difference between it and that of \(D_H\) is minimized via a MLP transmitter in the step 3. Finally, the supervised model of \(D_L\) is fine-tuned by the model that appends the pre-trained multi-task MLP of \(D_H\) in step 2 and the regularized encoder of \(D_L\) in step 3.}
\label{Fig: 01}
\end{figure*}

Figure \ref{Fig: 01} shows the overall framework of CLEIT. The training of CLEIT involves four steps: 1) learning an embedding of \(D_H\) from unlabeled data using variational autoencoder (VAE) \cite{vae}, 2) fine-turning the pre-trained embedding of \(D_H\) from the step 1 using a multi-layer perceptron (MLP) in the setting of multi-task supervised learning, 3) learning an embedding of \(D_L\) from unlabeled data using VAE along with training a MLP-based transmitter for the regularization of the \(D_L\) embedding by the \(D_H\) embedding, and 4) supervised learning of the final predictive model of \(D_L\) using an architecture that appends the pre-trained multi-task MLP from step 2 as well as the pre-trained VAE encoder and the transmitter of \(D_L\) from step 3. 
We denoted unlabeled \(D_H\) samples as \(\textbf{\emph{X}}_{H_{u}} = \{\textbf{\emph{x}}_{H_{u}}^{(i)}\}_{i=1}^{N_{H_u}}\) and labeled samples as \(\textbf{\emph{X}}_{H_{l}} =\{(\textbf{\emph{x}}_{H_{l}}^{(i)}, \textbf{\emph{y}}^{(i)})\}_{i=1}^{N_{H_l}}\), where \(N_{H_\cdot}\) stands for the number of samples in corresponding data sets. Furthermore, \(\textbf{\emph{z}}_{H_\cdot}\) is used to symbolize the latent vectors(variables) learned in different phases throughout the training. Samples from the \(D_L\) are similarly denoted.

\subsection{High-level domain \(D_H\) encoder training and fine-tuning}

For the pre-training of \(D_H\), we first constructed a VAE \cite{vae} to "warm" the encoder with standard input reconstruction task.
In the fine-tuning step, for a label space that has multiple tasks, we first appended several shared layers after the pre-trained encoder module, then for each task we append additional fully connected layer to make complete predictor per task. In our setting, the task is the anti-cancer sensitivity of a drug. We primarily consider a regression problem, thus the loss in use is the masked scaled-invariant mean squared error which allows the missing values in the multi-dimension label space, as defined below, 
\begin{align}
\mathcal{L}_{si-mse}(\textbf{\emph{y}}, \widehat{\textbf{\emph{y}}}) & = \frac{1}{N_{H_l}}\{\frac{1}{k^*}\left \| \left [ \textbf{\emph{y}}-\widehat{\textbf{\emph{y}}} \right ] \cdot \mathbb{I}_{ \left [ \textbf{\emph{y}}_{k_{j} \neq NA} \right ] } \right \|_2^2 \nonumber\\ & -\frac{1}{(k^{*})^2}(\left[\textbf{\emph{y}}-\widehat{\textbf{\emph{y}}} \right ] \cdot \mathbb{I}_{ \left [ \textbf{\emph{y}}_{k_{j} \neq NA} \right ] })^2\}
\end{align}
where \(\widehat{\textbf{\emph{y}}} \in \mathbb{R}^{k}\) is the predicted drug sensitivity score (vector), and \(\mathbb{I}_{[\textbf{\emph{y}}_{k_{j} \neq NA}]}\) is an indicator vector of length \(k\), that stands for the availability of  ground-truth label for this sample and \(k^*\) accordingly is the total number of tasks with ground truth score for this sample. For the fine-tuning of the encoder, we started with updating only appended predictor modules and then employed gradual unfreezing and layer-wise decayed learning rates on encoder updating. After the training of \(D_H\), the encoder is frozen to generate "ground-truth" latent representation of \(D_H\). In addition, the appended predictor modules (including shared and individual) can also be preserved to serve the initialized predictor in \(D_L\) to facilitate corresponding training process.

\subsection{Low-level domain \(D_L\) encoder and transmitter}
The pre-training of \(D_L\) is largely the same as the one used for the encoder training of \(D_H\). Different from \(D_H\), the encoder training in \(D_L\) is merely an intermediate step to improve the final predictive performance of tasks of interest. As shown in Figure \ref{Fig: 01}, we also used a VAE to pre-train the encoder with unlabeled data, followed by the fine-tuning of supervised modules. The key innovation lies in the information transmitter module between the hidden representations of two domains. As introduced earlier, we considered the latent representation generated by \(D_H\) encoder as "ground-truth" representation to which \(D_L\) encoder approximates. We utilized the additional transmitter module to explicitly bridge the asymmetrical encoding process between \(D_H\) and \(D_L\). Specifically, the transmitter is responsible for the minimization of the difference between the hidden representations of two domains. 

According to the formation of our training scheme, we are leveraging the stochastic encoder in the CLEIT framework. The latent variable \(\textbf{\emph{z}}\) can be seen as multi-variate Gaussian variable. Thus, proper distribution distance loss can be used to measure the difference between representations. Such cross-level information transmission loss between representations can then be used as additional regularization to guide the training of encoder of \(D_L\). We can then train the encoder in a multi-task setting, where the training loss is defined as weighted combination of VAE loss (\(\mathcal{L}_{VAE}\)) and cross-level information regularization loss (\(\mathcal{L}_{clr}\)) as shown below,
\begin{equation}
     \lambda \mathcal{L}_{clr}(\textbf{\emph{z}}_{L}, \textbf{\emph{z}}_{H}) + (1-\lambda) \mathcal{L}_{VAE}(\textbf{\emph{x}}_{L_{u}}, \widehat{\textbf{\emph{x}}_{L_u}}, \textbf{\emph{z}}_{L_u})
     \label{eq:clr}
\end{equation}
where $\lambda$ is a user-specified hyper-parameter to balance the loss terms, when $\lambda = 0$, the training does not use any information transmitted across domains. $F$ stands for additional transmission function, whose job is to explicitly transform \(\textbf{\emph{z}}_{L_u}\) (latent representation learned with auto-encoder training objective) to a representation that mimics the latent representation of \(D_H\). Thus, we have $\textbf{\emph{z}}_{L}=F\left (\textbf{\emph{z}}_{L_u} \right )$ which stands for the transmitted representation. In this work, a two layer fully connected MLP is used as the transmission function. 
We used contrastive loss for the transmitter. The contrastive loss is one of the newest losses being used in self-supervised learning framework \cite{contrastive}. We consider the  hidden representation pair of the same sample in difference domains as positive, and the contrastive loss is defined as, 
\begin{equation}
    \mathcal{L}_{ctr}(\textbf{\emph{z}}_{H}^{(i)}, \textbf{\emph{z}}_{L}^{(i)}) = -log \frac{exp(sim(\textbf{\emph{z}}_{H}^{(i)}, \textbf{\emph{z}}_{L}^{(i)}))}{\Pi}
\end{equation}
where 
\begin{align}
    \Pi & = \sum_{k=1}^{N}\mathbb{I}_{[k \neq i]}exp(sim(\textbf{\emph{z}}_{H}^{(i)}, \textbf{\emph{z}}_{L}^{(k)})) \nonumber \\
    & + \sum_{k=1}^{N}\mathbb{I}_{[k \neq i]}exp(sim(\textbf{\emph{z}}_{L}^{(i)}, \textbf{\emph{z}}_{H}^{(k)}))
\end{align}
and
\begin{equation}
    sim(\textbf{\emph{z}}_{H}^{(i)}, \textbf{\emph{z}}_{L}^{(k)}) = \frac{{\textbf{\emph{z}}_{H}^{(i)}}^{T} \textbf{\emph{z}}_{L}^{(k)}}{\left \| \textbf{\emph{z}}_{H}^{(i)} \right \| \left \| \textbf{\emph{z}}_{L}^{(k)} \right \| 
}
\end{equation}
\(\mathbb{I}_{[k \neq i]}\) is the indicator vector of condition \((k \neq i)\) and the cross-level information regularization loss computed as the average contrastive loss between all positive pairs within one batch.

\subsection{Low-level domain \(D_L\) supervised learning}
Since we are incorporating the cross-level information regularization loss into the encoder training, this requires us to leverage all the samples that have the features of both domains regardless of the label's availability in this encoder training phase. The pre-training of \(D_L\) encoder will terminate once stop condition is satisfied. The stopping condition can be maximum number of epochs or early stopping with validation dataset. 
In the supervised fine-tuning phase of \(D_L\) task, the inherited predictor modules from the training of \(D_H\) are appended after \(D_L\) encoder and transmitter to avoid the initial training on these modules. Like \(D_H\) encoder fine-tuning, we also adopted similar gradual unfreezing for the encoder and the transmitter module (when it is not an identity function) as well as layer-wise decayed learning rate. 

A detailed fine-tuning procedure can be found in (Procedure \ref{alg1}).
\algsetup{indent=2em}
\floatname{algorithm}{Procedure}
\begin{algorithm}[H]
\small
\caption{Fine-tuning of multi-task supervised learning}
\label{alg1}
\textbf{Input}: \(\{(\textbf{\emph{x}}_{G_l}^{(i)}, \textbf{\emph{y}}^{(i)})\}_{i=1}^{N_{G_l}}\)
\begin{algorithmic}[1]
\REQUIRE $m$, the batch size; $\alpha$, initial learning rate; $n_{f}$, number of epochs to keep encoder frozen; 
\(n_{uf}\), number of epochs to unfreeze one more layer; $\delta$, learning rate decay coefficient; $\theta$, all trainable parameters, begin with only regressor variables.
\FOR{epoch $=1$ to $n_{f}$}
\STATE Update \(\theta\) with \(\mathcal{L}_{si-mse}\)
\ENDFOR
\STATE epoch \(=0\)
\REPEAT
\IF{epoch \(\% n_{uf} == 0\) }
\STATE Expand \(\theta\) with highest frozen layer of the encoder
\STATE \(\alpha = \alpha * \delta\)
\ENDIF

\FOR{\(t=1\) to \(T\)} 
\STATE sample \(\{(\textbf{\emph{x}}_{G_{l}}^{(i)},  \textbf{\emph{y}}^{(i)})\}_{i=1}^{m}\) from \(\{(\textbf{\emph{x}}_{G_{l}}^{(i)},  \textbf{\emph{y}}^{(i)})\}_{i=1}^{N_{G_l}}\) (w/o. rep)
\STATE \(g_{\theta} \leftarrow \nabla_{\theta}  \mathcal{L}_{si-mse}\) 
\STATE \(\theta = \theta - \alpha \cdot g_{\theta}\)
\ENDFOR
\STATE \textbf{SHUFFLE} \(\{(\textbf{\emph{x}}_{G_{l}}^{(i)},  \textbf{\emph{y}}^{(i)})\}_{i=1}^{N_{G_l}}\)

\UNTIL{Stop Condition}

\end{algorithmic}
\end{algorithm}
 
\section{Experiments}
\subsubsection{Datasets}

We evaluate the performance of CLEIT on a real-world problem: predicting anti-cancer drug sensitivity given the mutation profile of cell lines. The mutation profile (oncology panel) has been implemented in clinic, but has weaker discriminative power for drug sensitivity prediction than the gene expression profile that is not a clinical standard yet. We collected and integrated data from several diverse resources: cancer cell line data from CCLE \cite{CCLE2}, pan-cancer data from Xena \cite{xena}, drug sensitivity data from  GDSC \cite{GDSC1}, and gene-gene interactions from STRING \cite {szklarczyk2019string}. CCLE includes 1270 and 1697 cancer cell line samples with the gene expression profile and the somatic mutation profile, respectively. The pan-cancer data sets include 9808 and 9093 tumor samples with the gene expression profile and the somatic mutation profile, respectively. 
All gene expression data are metricized by the standard transcripts per million base for each gene, with additional log transformation. For the somatic mutation data, we kept only non-silent ones then propagated the mutated genes in each sample on a STRING gene-gene interaction network using pyNBS \cite{huang2018pynbs}. Then, we only kept genes belonging to cancer gene consensus \cite{CGC}. 563 and 562 CGC genes are selected for the gene expression and the somatic mutation, respectively. Furthermore, we matched the omics data of CCLE cell lines against GDSC drug sensitivity score measured by Area Under Drug Response Curve (AUC). In total we assembled 575 CCLE cell lines with both mutation and gene expression, which are associated with the 265 anti-cancer drugs. 222 cell lines have only mutation information. These data are used as labeled data in our study. Finally, the matched unlabeled samples are also identified to facilitate the pre-training. The gene expression profile is considered as \(D_H\), while the mutation as \(D_L\). A brief summary of the pre-processed data are shown in Table~\ref{Tab: 01}.

\newcommand{\specialcell}[2][c]{%
  \begin{tabular}[#1]{@{}c@{}}#2\end{tabular}}
  
\begin{table}[h!]
\small
\centering
\scalebox{0.9}{
\begin{tabular}{@{}c|ccc@{}}

\toprule
\textbf{Category} & \specialcell{\textbf{Unlabeled}\\ (pre-training)} & \specialcell{\textbf{Labeled}\\(fine-tuning)} & \specialcell{\textbf{Labeled}\\(test)} \\ \midrule
\specialcell{\textbf{Gene Expression} \\ (\#cell lines)} & \begin{tabular}[c]{@{}c@{}}10503\end{tabular} & 575 & NA \\
\specialcell{\textbf{Somatic Mutation} \\ (\#cell lines)} & \begin{tabular}[c]{@{}c@{}}9993\end{tabular} & 575 & 222 \\
\specialcell{\textbf{Drug Sensitivity} \\ (\#cell line-drug pairs)} & NA & \begin{tabular}[c]{@{}c@{}}122189\end{tabular} & \begin{tabular}[c]{@{}c@{}}47432\end{tabular} \\ \bottomrule
\end{tabular}
}
\caption{Summary of Pre-processed Data}
\label{Tab: 01}
\end{table}

{\setlength\tabcolsep{3.7pt}\small
\begin{table*}
\centering

\begin{tabular}{lcccc}
\toprule
                                     & \multicolumn{2}{c}{\textbf{Drug-wise} }                   & \multicolumn{2}{c}{\textbf{Sample-wise} }                  \\
\multicolumn{1}{c}{\textbf{Method} } & Pearson                     & RMSE                        & Pearson                     & RMSE                         \\ 
\midrule
MLP (Mutation only)                  & 0.0414 $\pm 0.0017$         & 0.1679$\pm0.0021$           & 0.6441$\pm0.0040$           & 0.1659$\pm0.0027$            \\
VAE+MLP (Mutation only)              & 0.0628 $\pm 0.0031$         & 0.1544$\pm0.0041$           & 0.6630$\pm0.0107$           & 0.1491$\pm0.0064$            \\
DDC                                  & 0.0817$\pm0.0021$           & 0.1541$\pm0.0030$           & 0.6673$\pm0.0037$           & 0.1524$\pm0.0025$            \\
CORAL                                & 0.0732$\pm0.0033$           & 0.1579$\pm0.0021$           & 0.6624$\pm0.0021$           & 0.1507$\pm0.0015$            \\
DANN                                 & 0.0969$\pm0.0013$           & 0.1517$\pm0.0018$           & 0.6773$\pm0.0017$           & 0.1486$\pm0.0019$            \\
ADDA                                 & 0.0967$\pm0.0103$           & 0.1520$\pm0.0095$           & 0.6827$\pm0.0093$           & 0.1488$\pm0.0088$            \\
DSN                                  & 0.1413$\pm0.0041$           & 0.1419$\pm0.0026$           & 0.6922$\pm0.0024$           & 0.1322$\pm0.0026$            \\
CLEIT (w/o transmitter)              & 0.1456$\pm0.0044$           & 0.1343$\pm0.0077$           & 0.6930$\pm0.0064$           & 0.1238$\pm0.0080$            \\
CLEIT (MMD)                          & 0.1462$\pm0.0049$           & 0.1384$\pm0.0025$           & 0.6943$\pm0.0041$           & 0.1351$\pm0.0027$            \\
CLEIT (WGAN)                         & 0.1223$\pm0.0095$           & 0.1504$\pm0.0080$           & 0.6843$\pm0.0122$           & 0.1428$\pm0.0103$            \\
\textbf{CLEIT (Contrastive)}         & \textbf{0.1630}$\pm0.0035$  & \textbf{0.1209}$\pm0.0039$  & \textbf{0.7171}$\pm0.0035$  & \textbf{0.1158}$\pm0.0085$   \\ 
\bottomrule
\end{tabular}
\caption{Evaluation Results on Test Data. CLEIT significantly outperforms all state-of-the-art base-line models.}
\label{Tab: 02}
\end{table*}
}

\subsubsection{Experiment Set-up}
We evaluated the performance of CLEIT by the task of predicting drug sensitivity on hold-out test over the labeled mutation-only test data. To be noted, both the prediction and ground truth are in the format of a matrix. Each row represents the sensitivity scores of a particular sample (cell line) against all drugs, and each column stands for the sensitivity scores of a specific drug against all tested cell lines. Thus the evaluation needs to be done by both sample-wise (per sample) and drug-wise (per drug). The evaluation metrics in use include the Pearson correlation and RMSE (root mean squared error). In addition, because of the incompleteness of the ground truth matrix, the prediction entries without a ground truth sensitivity score are filtered out in the calculation of each evaluation metric.

\subsubsection{Training procedure of CLEIT}
The training procedure of CLEIT is as follows. In the \(D_{H}\) pre-training, we employed early stopping by tracking the VAE loss performance of labeled \(X_{H_l}\). While for the fine-tuning of \(D_{H}\), we set the maximum number of training epochs according to cross-validation experiment results. For the pre-training of \(D_{L}\), we used the labeled \((X_{H_l}, X_{L_l})\) as validation set to apply early stopping strategy. Finally, we split \(X_{L_l}\) into 90\% training and 10\% validation set for the fine-tuning of \(D_{L}\), where the validation set is used for early stopping. The final trained model are used to make predictions on labeled mutation-only test set. We repeat the \(D_{L}\) fine-tuning 3 times.

\subsubsection{Baseline models}
We compared CLEIT with the following base-line models: MLP without and with the VAE pre-training for \(D_L\) as well as several of the most popular domain adaptation algorithms that are used to transfer the knowledge learned from \(D_H\) to \(D_L\). They include Deep Domain Confusion (DDC) network \cite{DCC2014}, Correlation Alignment (CORAL) \cite{coral2015}, Domain Adversarial Neural Network (DANN)\cite{dann2016}, Adversarial Domain Adaptation Network (ADDA) \cite{adda2017} and Domain Separation Network (DSN) \cite{dsn}. Specifically, DDC, CORAL, DANN, and ADDA only made use of the labeled data, while DSN utilized both the unlabeled and labeled data. For domain adversarial loss in DSN we employed the MMD variant for the stability of training. 

To evaluate the contribution of different components in CLEIT, we performed ablation studies by 1) removing the transmitter, 2) change the loss function to Maximum Mean Discrepancy (MMD) loss \cite{MMD} and Wasserstein-GAN (WGAN) \cite{wgan}. Evaluation results on test data for CLEIT, base-line models, and ablation studies are summaries in Table \ref{Tab: 02}.
\subsubsection{Hyperparameters of CLEIT}
Neural network-related models are constructed using TensorFlow 2.1. The detailed architecture of CLEIT and hyperparameters are listed in Table \ref{Tab: 03},

\begin{table}[h!]
\centering
\scalebox{0.8}{
\begin{tabular}{@{}c|c@{}}

\toprule
\textbf{Components} & \specialcell{\textbf{Description}}  \\ \midrule
\specialcell{\textbf{\(z\) Dimension}} & \specialcell{128\\} \\ \midrule
\specialcell{\textbf{Encoders}} & \specialcell{layers: [512, 256, 128]\\middle layers: SELU + layer norm + dropout\\last layer: layer norm} \\ \midrule
\specialcell{\textbf{Decoders}} & \specialcell{layers: [128, 256, 512]\\middle layers: SELU + layer norm + dropout\\last layer: linear} \\ \midrule
\specialcell{\textbf{Regressors}} & \specialcell{shared layers: [128, 128]\\individual module layers: [64, 16, 1]\\ output activation: sigmoid} \\ \midrule
\specialcell{\textbf{Transmitter}} & \specialcell{layers: [128, 128]\\middle layer: SELU\\last layer: layer norm} \\ \midrule
\specialcell{\textbf{Pre-training}} & \specialcell{batch size: 64\\learning rate: 5e-3\\optimizer: Adamax\\\(\lambda\) (\(D_{L}\), equation (\ref{eq:clr})):0.8} \\ \midrule
\specialcell{\textbf{Fine-tuning}} & \specialcell{batch size: 64\\learning rate: 1e-4\\optimizer: Adamax\\decay coefficient: 0.8} \\ 

\bottomrule
\end{tabular}
}
\caption{Details of CLEIT architecture and hyperparameters}
\label{Tab: 03}

\end{table}
\section{Results and Discussion}
\begin{figure*}[t!]
    \centering
    \includegraphics[width=1.0\textwidth]{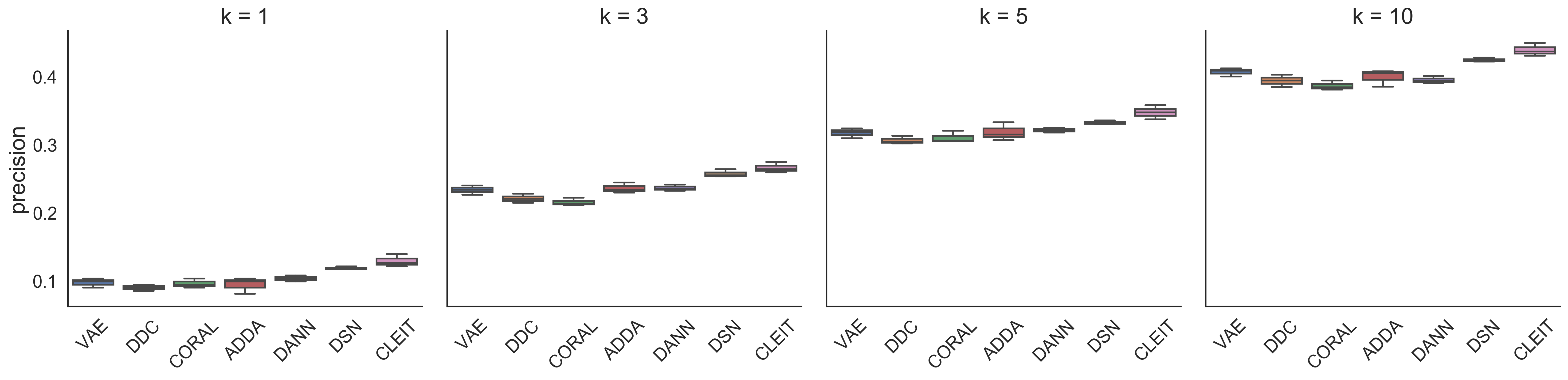}
    \caption{Top K Precision on Mutation-only Test Dataset}
    \label{Fig: 02}
\end{figure*}
\subsubsection{Comparison with state-of-the-art models.}
The results for both drug-wise and sample-wise evaluation are shown in Table \ref{Tab: 02}. As seen in Table \ref{Tab: 02}, all base-line domain adaptation models outperform the simple models with MLP only and VAE plus MLP in the drug-wise setting. It implies that \(D_L\) will benefit from the knowledge transfer from \(D_H\). Furthermore, CLEIT models significantly outperform all other models in consideration (\textit{t}-test \textit{p}-value $<$ 0.05). The best performed model is the CLEIT that uses the contrastive loss. Compared with the best performed state-of-the-art model (DSN), the accuracy of CLEIT, when measured by Pearson's correlation, improves 13.3\% and 3.5\% for the drug-wise and the sample-wise test, respectively. 

\subsubsection{Ablation studies.}
In addition, the pre-training is helpful in improving the model performance, as suggested by the MLP models in Table \ref{Tab: 02}. CLEIT models that incorporate MLP-transmission function show significantly better performance than the one without, suggesting that the transmission function plays a role in CLEIT. Choice of the loss function in the information transmission is also important. It is clear that contrastive loss performs better than MMD and WGAN. It is noted that MMD is used in DSN. When CLEIT uses MMD as the loss function to measure the domain discrepancy, the major difference between CLEIT and MMD is that CLEIT treats the information transmission between two domains asymmetrical, while DSN considers domain adaptation symmetrical. The results in Table \ref{Tab: 02} show that CLEIT-MMD outperforms DSN in both drug-wise and sample-wise settings. It indicates that the explicit modeling of hierarchical organization of \(D_L\) and \(D_H\) is important.   

\subsubsection{Prediction of top-ranked cell-line specific anti-cancer therapies.}
Furthermore, CLEIT can be used to predict the best therapy for a new patient using only mutation data for precision medicine. We compared the performance of different methods with the precision of top-\(k\) (\(k=1,3,5,10\)) predictions ranked by the AUC scores, which is defined as the ratio of drugs with top-\(k\) smallest predicted scores per cell line among the drugs with top-\(k\) ground-truth scores. Mutation only test results can be found below in Figure \ref{Fig: 02}. Clearly, CLEIT model also outperforms other models in this scenario. Compared with the second best performed model DSN, CLEIT improves the performance by approximate 5\% when \(k\) = 5.  

\section{Conclusion}

In this paper, we propose a novel machine learning framework CLEIT for the predictive modeling of genotype-phenotype associations by explicitly modeling the asymmetric cross-level information transmission in a biological system. Using the anti-cancer drug sensitivity prediction with only mutation data as a benchmark, CLEIT clearly outperforms existing methods and demonstrates its potential in precision medicine. Nevertheless, the performance of CLEIT could be further improved by incorporating domain knowledge. For example, an autoencoder module that can model gene-gene interactions and biological pathways will be greatly helpful. Under the framework of CLEIT, it is not difficult to integrate other omics data such as epigenomics and proteomics. They may further improve the performance of CLEIT.    

\section{Ethic Statement}
This research addresses a fundamental problem in biology and ecology and will benefit broad scientific communities in both basic and translational studies. One of the immediate application of this research is in precision medicine. The proposed CLEIT can be used to inform the most effective therapy for a cancer patient based on her/his somatic mutation profile that is often included in the diagnosis. Caution should be taken when applying this research to precision medicine. The same as any machine learning techniques in the biomedical application, the outcome from this research only helps for the decision making but does not provide the final clinical decision, which should be made by a health professional. 

\bigskip

\bibliography{cleit.bib}

\end{document}